\documentclass[10pt,twocolumn,letterpaper]{article}

\usepackage{authblk}
\usepackage[pagenumbers]{wacv} % To force page numbers, e.g. for an arXiv version
\usepackage[accsupp]{axessibility}
\usepackage{graphicx}
\usepackage{amsmath}
\usepackage{amssymb}
\usepackage{booktabs}
\usepackage{multirow}
\usepackage{mathtools}
\usepackage{xspace}

\newcommand{\mbh}{\mathbf{h}}

\newcommand{\mbm}{\mathbf{m}}

\newcommand{\mbp}{\mathbf{p}}

\newcommand{\mbs}{\mathbf{s}}

\newcommand{\mbC}{\mathbf{C}}

\newcommand{\mbF}{\mathbf{F}}

\newcommand{\mbI}{\mathbf{I}}
\newcommand{\mbJ}{\mathbf{J}}

\newcommand{\mbM}{\mathbf{M}}
\newcommand{\mbN}{\mathbf{N}}

\newcommand{\mbT}{\mathbf{T}}

\newcommand{\ignore}[1]{}

\makeatletter
\DeclareRobustCommand\onedot{\futurelet\@let@token\@onedot}
\def\@onedot{\ifx\@let@token.\else.\null\fi\xspace}
\def\ie{{i.e}\onedot} %\def\Ie{{I.e}\onedot}

\def\etal{{et al}\onedot}
\makeatother

\newcommand\blfootnote[1]{%
  \begingroup
  \renewcommand\thefootnote{}\footnote{#1}%
  \addtocounter{footnote}{-1}%
  \endgroup
}
\usepackage[pagebackref,breaklinks,colorlinks]{hyperref}

\usepackage[capitalize]{cleveref}
\crefname{section}{Sec.}{Secs.}
\Crefname{section}{Section}{Sections}
\Crefname{table}{Table}{Tables}
\crefname{table}{Tab.}{Tabs.}

\def\wacvPaperID{1861} % *** Enter the WACV Paper ID here
\def\confName{WACV}
\def\confYear{2024}

\begin{document}

\title{3D Reconstruction of Interacting Multi-Person in Clothing from a Single Image}

% \and
\author{Junuk Cha\textsuperscript{1} \qquad
Hansol Lee\textsuperscript{1,2\dag} \qquad
Jaewon Kim\textsuperscript{1} \qquad
Nhat Nguyen Bao Truong\textsuperscript{1,3\ddag} \qquad

Jae Shin Yoon\textsuperscript{4*} \qquad
Seungryul Baek\textsuperscript{1*}
}
\affil{\textsuperscript{1}UNIST \qquad \textsuperscript{2}KIST \qquad \textsuperscript{3}KAIST \qquad \textsuperscript{4}Adobe Research}

\twocolumn[{
\maketitle
\begin{center}
    \captionsetup{type=figure}
    \vspace{-6mm}
    \includegraphics[width=0.99\textwidth]{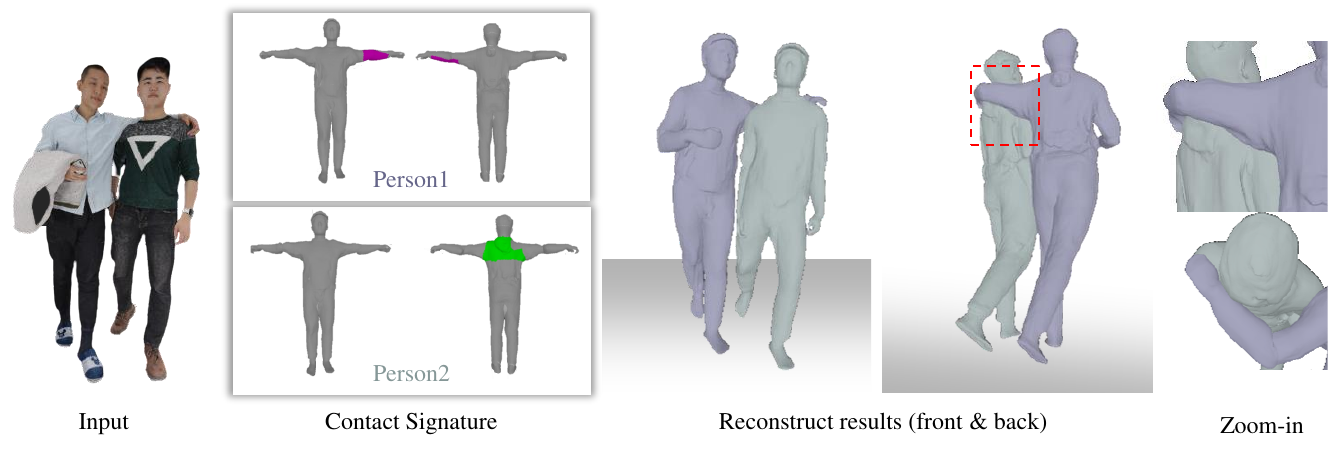}
    \vspace{-3mm}
    \captionof{figure}{Given a single RGB image of interacting multiple people with occlusion, we reconstruct their complete geometry. Our method estimates the contact signature between people, which provides a strong pose refinement cue in 3D to prevent penetration.}
    \label{fig:teaser}
\end{center}
}]

\blfootnote{This research was conducted when Hansol Lee was a graduate student (Master candidate) at UNIST$\dag$, and when Mr. Bao was undergraduate interns at UNIST$\ddag$. co-last authors$*$.}
%%%%%%%%% ABSTRACT
\begin{abstract}
This paper introduces a novel pipeline to reconstruct the geometry of interacting multi-person in clothing on a globally coherent scene space from a single image.
The main challenge arises from the occlusion: a part of a human body is not visible from a single view due to the occlusion by others or the self, which introduces missing geometry and physical implausibility (e.g., penetration). 
We overcome this challenge by utilizing two human priors for complete 3D geometry and surface contacts.   
For the geometry prior, an encoder learns to regress the image of a person with missing body parts to the latent vectors; a decoder decodes these vectors to produce 3D features of the associated geometry; and an implicit network combines these features with a surface normal map to reconstruct a complete and detailed 3D humans. 
For the contact prior, we develop an image-space contact detector that outputs a probability distribution of surface contacts between people in 3D.
We use these priors to globally refine the body poses, enabling the penetration-free and accurate reconstruction of interacting multi-person in clothing on the scene space.
The results demonstrate that our method is complete, globally coherent, and physically plausible compared to existing methods.
\end{abstract}

%%%%%%%%% BODY TEXT
\section{Introduction}
Knowing people's underlying geometry makes it possible to transport the scenes of human interaction to a virtual space, which is the key component for AR/VR applications such as authentic telepresence and view-consistent rendering of people for internet broadcasting. Such multi-person geometry has been often captured by a specialized infrastructure, e.g., a multi-camera system~\cite{saito2019pifu,zheng2021deepmulticap}, for two reasons: 1) it maximizes the visibility of a scene, which enables the 3D reconstruction of people without missing parts; and 2) multi-camera calibration makes it possible to understand the scene-space 3D body poses. In this paper, we remove those requirements of an expensive system by introducing a novel pipeline that can capture the scene-space geometry of interacting multi-person using only a single image as described in Fig.~\ref{fig:teaser}.

\begin{figure*}[t]
\centering
\vspace{-5mm}
\includegraphics[width=0.99\textwidth]{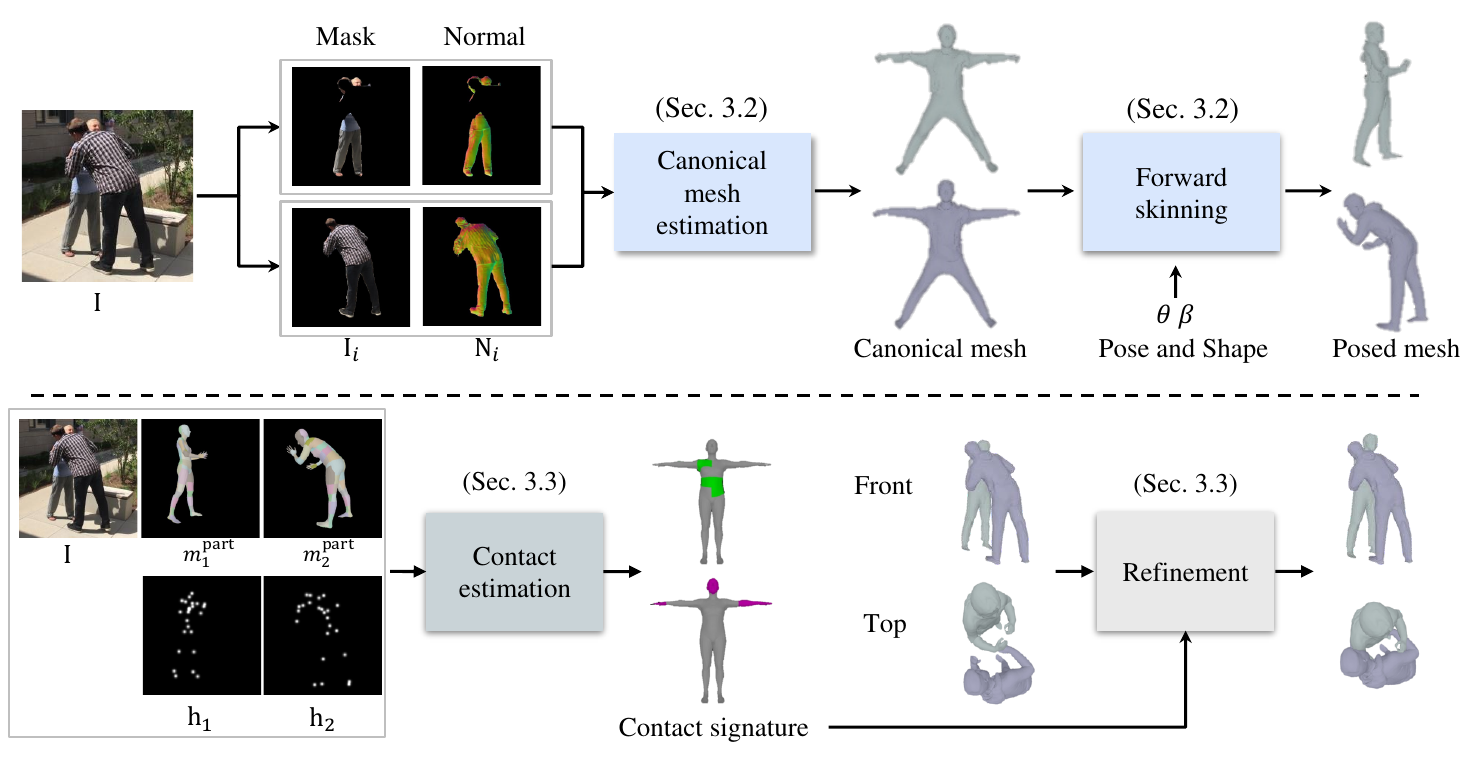}
\vspace{-5mm}
\caption{\textbf{System overview}. Given an image of interacting people, we aim to reconstruct multi-person geometry. Our pipeline is composed of three stages. (Top): In the generation stage, we extract coarse meshes for the multi-person from the input image. (Bottom): In the contact estimation stage, we detect the region of contact between individuals in the image. Finally, in the contact-based refinement stage, we generate detailed multi-person geometry by leveraging the information obtained from the previous two stages.} 
% \vspace{-3mm}
\label{fig:pipeline}
\end{figure*}

Monocular multi-person 3D reconstruction is a highly challenging problem due to self- and inter-occlusion: a significant amount of missing body parts in an image is unavoidable from a single-view perspective during multi-person interaction. Such incomplete information in 2D, in turn, affects the 3D reconstruction with missing geometry. While existing methods~\cite{cha2022multi,kanazawa2018end,kocabas2021pare,kolotouros2019learning} have utilized a 3D body model, e.g., SMPL~\cite{loper2015smpl}, to capture a complete human, they are limited to expressing unclothed human appearance; and the reconstruction of the multi-person is often physically implausible since it involves unrealistic penetration artifacts and scene-space positional misalignment (i.e., a person's position in world coordinates is unusually distant from others compared to their actual physical distance, as shown in Fig.~\ref{fig:pipeline}, bottom right).

To address this challenge, we propose to use human priors for complete 3D geometry and surface contacts. For the complete geometry, we effectively utilize a 3D generative model (e.g., gDNA~\cite{chen2022gdna}) which is designed to generate 3D features from a latent vector, for a coarse and looking-plausible human geometry. We upgrade this 3D geometric coarse prior to 3D geometric detail prior by developing an implicit network that combines the 3D features~\cite{chen2022gdna} and a surface normal map. It predicts the 3D geometry augmented in its style, shape, and details (e.g., wrinkles). We finally glue the different modalities between the image and the 3D model by designing an image regression network that can predict the appearance vectors whose distribution matches the latent space of the 3D generative model~\cite{chen2022gdna}. For understanding the physical state of multiple people in 3D, we develop a data-driven prior for surface contact map---a probability distribution of 3D surface contacts of each body part across people. A neural network learns to predict such probability maps from the images of interacting multi-person, which provides strong guidance for accurate pose refinement in scene space. In particular, it is designed to prevent surface penetration; and to enforce accurate scene-space localization, e.g., the contact around the torso of two people in Fig.~\ref{fig:pipeline} provides a strong cue to decide their ordinal depth relation.

Using these two priors, we introduce a new design of the monocular multi-human capturing system.
Given an image of multiple people, we first predict the detailed geometry for each person in the canonical body poses using our 3D geometric detail prior.
The coarse scene-space 3D body location for each person are estimated by solving Perspective-n-Point (PnP) between the predicted 3D keypoints and a 3D body model (SMPL~\cite{loper2015smpl}).
The multi-person poses are then refined by optimizing their global relation based on the predicted contact maps and penetration loss. 

The experimental results demonstrate the strength of our method in geometric completeness with the upgraded 3D geometric detail priors; in global coherence thanks to the contact priors; and in physical plausibility with minimal penetration with the effective refinement pipeline.

Our contribution includes (1) a novel pipeline that realizes 3D reconstruction of interacting multi-person using only a single image with an effective global pose refinement; (2) the innovation to upgrade existing geometric priors that enable a complete and detailed 3D human reconstruction from the image of a partial body; and (3) an effective scene-space pose refinement framework guided by surface contact priors, which enforces the physical plausibility. 

\section{Related Work}
% We summarize the literature that introduced the methods for 3D human reconstruction using a single image or video.

\noindent\textbf{3D Clothed Single Person Reconstruction.}
%Many methods have been developed to reconstruct the human mesh from single-view input. Some methods~\cite{cha2022multi,kanazawa2018end,kocabas2021pare,kolotouros2019learning,sun2021monocular} regress SMPL parameters~\cite{loper2015smpl} from RGB image. However, they can not express their clothing, shoes, and hair. To express more details of humans, some methods~\cite{saito2019pifu, saito2020pifuhd, xiu2022icon,xiu2022econ} have been proposed. However, they focus on reconstructing only one human mesh.  Thus, their performance is poor if there is an occlusion in the single-view input image.
Many existing methods have explored the 3D reconstruction of human meshes from a single-view image. Some works~\cite{kanazawa2018end,kocabas2021pare,kolotouros2019learning,sun2021monocular} predict SMPL parameters~\cite{loper2015smpl} from RGB images of a person with occlusion to reconstruct a complete human mesh. However, these works cannot express the shape beyond the naked body. 

Some explicit-based approaches~\cite{alldieck2019learning,alldieck2019tex2shape,lazova2019360,zhu2019detailed} solved this problem by estimating the offsets at each vertex of the body mesh to represent the humans in more detail. Those explicit approaches have demonstrated sub-optimal performance when dealing with a single-view input, mainly because of the fixed 3D mesh topology~\cite{loper2015smpl}.

To overcome the limitation of the fixed topology constrained by explicit 3D mesh models, some existing methods have utilized an implicit signed distance field (SDF)~\cite{xiu2022icon,zheng2021deepmulticap,zheng2021pamir,alldieck2022photorealistic} and occupancy field~\cite{zheng2019deephuman,he2020geopifu,huang2020arch,saito2019pifu,saito2020pifuhd,zins2021data,tiwari2021neural, chen2021snarf, saito2021scanimate, wang2021metaavatar, kim2022laplacianfusion,he2021arch++,he2020geo,xiu2023econ}. To generate a high-resolution mesh from a coarse one in the canonical space, Neural-GIF~\cite{tiwari2021neural} uses a backward mapping network and a displacement network. LaplacianFusion~\cite{kim2022laplacianfusion} employed Laplacian coordinates extracted from the mesh to generate the detailed surface. While the above methods demonstrate promising reconstruction results, those methods often suffer from missing information in 3D when the input images of humans involve occlusion.

\noindent\textbf{3D Clothed Multi-Person Reconstruction.} 
The multi-person 3D reconstruction is often considered as a special case of a single person with occlusion where several methods~\cite{cha2022multi,choi2022learning,hassan2019resolving,zhang2021body,sun2021monocular,jiang2020coherent,zanfir2018deep,zanfir2018monocular,sun2022putting} separately handle each individual from the scenes of human interaction. However, the focus of these methods is lying on the reconstruction of unclothed human models (e.g., SMPL~\cite{loper2015smpl}) that cannot represent the details of human geometry such as clothing, hair, and shoes.

While DeepMultiCap~\cite{zheng2021deepmulticap} utilizes the attention module and temporal fusion to capture high-quality 3D models of interacting people in clothing, they require multiview images for the best quality of 3D reconstruction in scene space.

Mustafa~\etal~\cite{mustafa2021multi} introduced an implicit approach for 3D reconstruction of multiple people with 6DOF spatial location estimation to accurately estimate the position of people from global scene space. However, the results are limited to expressing coarse geometry and the method often fails when it faces people with heavy occlusion.

\noindent\textbf{Surface Contact Estimation.}
Surface contacts provide a strong cue to understand the physical state between two different surfaces. Thus, there has been a growing interest in modeling such surface contacts in the process of 3D human reconstruction for human-to-object~\cite{bhatnagar2022behave,xie2022chore}, human-to-scene~\cite{huang2022capturing,shimada2022hulc}, and human-to-human~\cite{fieraru2020three}. Fieraru~\etal~\cite{fieraru2020three} introduced a representation of part-based contact signatures at a 3D body model~\cite{loper2015smpl} to predict accurate 3D body poses from the scenes of multiple people. However, the above methods do not consider the surface geometry of clothed humans and are brittle under severe occlusions.

\section{Method}
The overall pipeline is illustrated in Fig.~\ref{fig:pipeline}. Our method consists of $3$ main stages which will be explained in detail in the remaining sections.

\subsection{Pre-processing}
\noindent \textbf{Human Segmentation.} We segment each human instance from an image $\mbI$ by using an existing method~\cite{li2020self}. From this, we could obtain the $i$-th person's segmentation mask $\mbm_i$. 

\noindent \textbf{Normal Map.}
We use an off-the-shelf normal detection method~\cite{saito2020pifuhd} to estimate the normal map $\mbN_{i}$ from $\mbI_{i}$. While it generates front and back normal maps from $\mbI_{i}$, we used only the front normal map.  

\noindent \textbf{SMPL Parameters.} We first initialize the 2D and 3D joints ($\mbJ^\text{2D}$, $\mbJ^\text{3D}$) from a single image by using an existing monocular keypoints prediction method~\cite{cha2022multi}. The inverse kinematic method~\cite{li2021hybrik} is applied to calculate the SMPL~\cite{loper2015smpl} pose parameter $\boldsymbol{\theta}$ and shape  parameter $\boldsymbol{\beta}$ from the estimated 3D joints $\mbJ^\text{3D}$.

\noindent \textbf{Camera Parameter and 3D Translation.} We estimate camera parameters $\mbC_\text{param}$, i.e., scale $s$, x-translation $t_x$, y-translation $t_y$, via the singular value decomposition (SVD) method to align SMPL mesh $\mbM_\text{smpl}$ from the image space: 
\begin{eqnarray}
    \mbJ^\text{2D}_{\text{smpl}}
    \begin{bmatrix}
    s \\
    t_x \\
    t_y \\
    \end{bmatrix}
    =
    \mbJ^\text{2D}.
\end{eqnarray}
% where
% \begin{eqnarray}
% \mbJ^\text{2D}_\text{smpl} = \begin{bmatrix}
%     \mbJ^\text{2D}_{\text{smpl}}(1,x) & 1 & 0 \\
%     & \vdots & \\
%     \mbJ^\text{2D}_{\text{smpl}}(J,x) & 1 & 0 \\
%     \mbJ^\text{2D}_{\text{smpl}}(1,y) & 0 & 1 \\
%     & \vdots & \\
%     \mbJ^\text{2D}_{\text{smpl}}(J,y) & 0 & 1 \\
%     \end{bmatrix},
%     \mbJ^\text{2D}=\begin{bmatrix}
%     \mbJ^\text{2D}(1,x) \\
%     \vdots \\
%     \mbJ^\text{2D}(J,x) \\
%     \mbJ^\text{2D}(1,y) \\
%     \vdots \\
%     \mbJ^\text{2D}(J,y) \\
%     \end{bmatrix}.
% \end{eqnarray}
% where $J$ denotes the number of joints. The SMPL 3D joints $\mbJ^\text{3D}_\text{smpl}$ are differentiably obtained from the mesh $\mbM_\text{smpl}$ using the SMPL skeletal regressor provided by~\cite{loper2015smpl}. The SMPL 2D joints $\mbJ^\text{2D}_\text{smpl}$ are obtained from the 3D joint $\mbJ^\text{3D}_\text{smpl}$ based on the orthogonal projection. We calculate 3D location $\mbT^\text{3D}$ with $\mbC_\text{param}$ and $\mbJ^\text{3D}_\text{smpl}$ similar to \cite{sun2021monocular,sun2022putting}.
The SMPL 2D joints $\mbJ^\text{2D}_\text{smpl}$ are obtained from the SMPL 3D joints $\mbJ^\text{3D}_\text{smpl}$ based on the orthogonal projection and $\mbJ^\text{3D}_\text{smpl}$ are differentiably obtained from the mesh $\mbM_\text{smpl}$ using the SMPL skeletal regressor provided by~\cite{loper2015smpl}. Additionally, we calculate 3D location $\mbT^\text{3D}$ with $\mbC_\text{param}$ and $\mbJ^\text{3D}_\text{smpl}$ similar to \cite{sun2021monocular,sun2022putting}.

\noindent \textbf{Semantic Part Segmentation and 2D Keypoint Heatmap.} We generate a semantic part segmentation mask $\mbm^\text{Part}_i$ by differentiably rendering 75 regions of SMPL 3D mesh $\mbM_\text{smpl}(\boldsymbol{\theta}, \boldsymbol{\beta})$ with different colors. We generate 2D keypoint heatmap $\mbh_i$ from 2D joints $\mbJ^\text{2D}$, by drawing $\mbJ^\text{2D}$ Gaussian in the positions of 2D joints.

\begin{figure*}[t]
    \centering
    \vspace{-3mm}
    \includegraphics[width=0.99\linewidth]{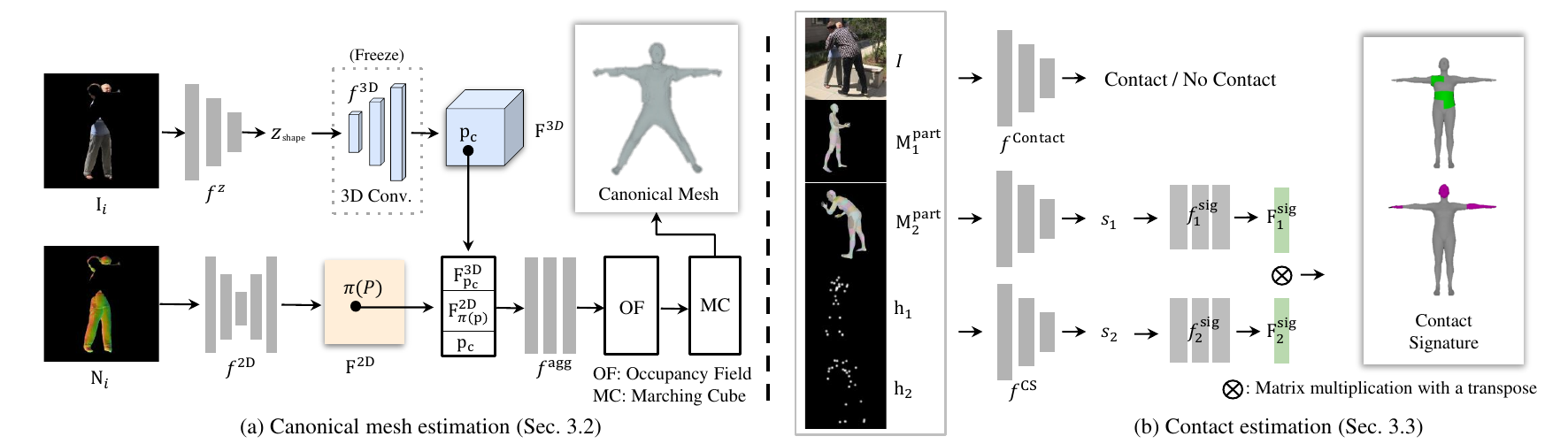}
    \caption{The details of the pipeline for each module in our system. }
     % (a) The canonical mesh estimation module in the Generation stage. (b) Contact estimation module in the contact estimation stage.
    \label{fig:pipeline_datail}
    \vspace{-5mm}
\end{figure*}

\subsection{Detailed Clothed Human Generation} 
In this section, our goal is to generate the human surface, reflecting details such as clothes and hair. As the direct reconstruction of holistic 3D meshes of detailed humans is non-trivial, we divide the skinning methods into two disjoint steps: First, we estimate the detailed 3D human mesh in the canonical space. Second, we transform the canonical 3D mesh into the posed mesh $\mbM$ corresponding to SMPL~\cite{loper2015smpl} pose $\boldsymbol{\theta}$ and shape $\boldsymbol{\beta}$ parameters using the forward skinning module. In the remainder of this sub-section, we detailed each sub-module.

\noindent\textbf{Canonical Mesh Estimation Module.} Our canonical mesh estimation module is designed to generate the canonical mesh $\tilde{\mbM}$. This module is composed of four sub-networks: a latent feature extractor $f^z$, a 3D voxel feature generator $f^\text{3D}$, a 2D image-aligned feature extractor $f^\text{2D}$ and an aggregation network $f^\text{agg}$ as shown in the left side of Fig.~\ref{fig:pipeline_datail}. 

The latent feature extractor $f^z$ extracts the latent vector $z_\text{shape}$ that reflects global feature of human styles from an input image $\mbI$. The 3D voxel feature generator $f^\text{3D}$ converts it into the feature vector $\mbF^\text{3D}$. The 2D image-aligned feature extractor $f^\text{2D}$ extracts the 2D image-aligned feature $\mbF^\text{2D}$ from a normal map $\mbN$. 

For any 3D point $\mbp$, we transform it to $\mbp_\text{c}$ in canonical space~\cite{chen2022gdna}. We obtain the associated canonical features $\mbF^\text{3D}_{\mbp_\text{c}}$ by tri-linear interpolation to the $\mbF^\text{3D}$ according to $\mbp_\text{c}$. $\pi(\mbp)$ is the perspective camera projection from the 3D point $\mbp$ to the image plane and the local 2D feature $\mbF^\text{2D}_{\pi(\mbp)}$ is obtained from the $\mbF^\text{2D}$ by bi-linear interpolation at the projected pixel coordinates $\pi(\mbp)$. Then, we concatenate $\mbF^\text{3D}_{\mbp_\text{c}}$, $\mbF^\text{2D}_{\pi(\mbp)}$ and $\mbp_\text{c}$ to make the geometry- and image-aligned feature. Finally, it is fed to the aggregation network $f^\text{agg}$ to obtain the high fidelity occupancy field $o(\mbp)$. Similar to~\cite{saito2019pifu}, the occupancy field $o({\mbp})\in[0,1]$ indicates whether ${\mbp}$ is inside mesh surface or not. Using the Marching Cube algorithm~\cite{lorensen1987marching}, we extract the final canonical mesh $\tilde{\mbM}$ from obtained occupancy fields $\{o({\mbp})\}$.

We initialize our $f^\text{3D}$ using the pre-trained model in~\cite{chen2022gdna}, while further training $f^z$, $f^\text{2D}$ and $f^\text{agg}$ to upgrade their method in a way that (1) infers the style code from the 2D images; (2) extracts geometric details from the normal map $\mbN$; and (3) combines information from both normal map and RGB images. 

To train $f^{z}$, we use the mean squared error (MSE) loss as follows:
\begin{eqnarray}
    L_z(f^z) = ||z_\text{shape}-z^\text{gDNA}_\text{shape}||^2_2,
\end{eqnarray}
where $z^\text{gDNA}_\text{shape}$ denotes the fitted latent code by gDNA~\cite{chen2022gdna}.
To train $f^\text{2D}$ and $f^\text{agg}$, we use the binary cross entropy (BCE) loss~\cite{zheng2019deephuman},
\begin{eqnarray}
    L_{o} &=& \sum_{\mbp\in S} \gamma \cdot o^\text{GT}(\mbp)\cdot \log( o(\mbp)) \nonumber\\
          &+& (1-\gamma) (1-o^\text{GT}(\mbp))\log(1-o(\mbp)),
\end{eqnarray}
where $S$ is the set of sampled points and $\gamma$ denotes the ratio of points outside the surface in $S$, which helps to train the network more stably. $o^\text{GT}(\mbp)$ is the ground-truth occupancy at the sampled point $\mbp$.

\noindent \textbf{Forward Skinning.}  We use the pre-trained network~\cite{chen2022gdna} to deform canonical mesh $\tilde{\mbM}$ towards the posed mesh $\mbM$ with SMPL~\cite{loper2015smpl} pose $\theta$ and shape $\beta$ parameters. It generates matrix $W \in \mathbb{R}^{N_\text{v} \times 24}$ that transforms $N_v$ canonical mesh vertices into skinning weight fields for $24$ bones, conditioned on $z_\text{shape}$ and $\beta$. As a result, given target $\theta$ and $W$, canonical mesh vertices $v$ on $\tilde{\mbM}$ can be transformed to the posed space $v'$ on the posed mesh $\mbM$. 

\subsection{Contact Map-based Refinement}
\noindent \textbf{Contact Estimation Stage.}
To predict contact estimation from images, we design a contact estimation network that takes an image, semantic part segmentation and 2D keypoint heatmap as inputs; and outputs contact signatures~\cite{fieraru2020three}, i.e., a representation that indicates whether there exist surface contacts between the body parts of two people from the 3D mesh model, as shown in Fig.~\ref{fig:pipeline}. We concatenate $\mbI$, $\mbm^\text{Part}_1$, $\mbm^\text{Part}_2$, $\mbh_1$, and $\mbh_2$ to feed them to the contact discriminator $f^\text{Contact}$ and the contact segmentation estimator $f^\text{CS}$.

The contact discriminator $f^\text{Contact}$ is designed to classify whether there exist contacts between two interacting persons. The contact segmentation estimator $f^\text{CS}$ is designed to predict the contact segmentation $\mbs_1\in[0,1]^{75\times 1}$ and $\mbs_2\in[0,1]^{75\times 1}$ for two persons. These are $75$-dimensional vectors whose values indicate whether each region out of the overall 75 regions is contacted or not. These two intermediate vectors are further fed to MLP layers $f^\text{sig}$ whose output vectors are used to infer the final contact signature $\mbC \in [0,1]^{75\times75}$ based on their dot products followed by a nonlinearity, i.e., sigmoid.

The overview of the network for this stage is shown on the right side of Fig.~\ref{fig:pipeline_datail}. To better distinguish the contact signature features of two interacting people, we employ the two-path architecture for each person and jointly estimating segmentation and signature is synergistic as described in ~\cite{fieraru2020three}.
% The overview of the network for this stage is shown on the right side of Fig.~\ref{fig:pipeline_datail}. To better distinguish the feature representation of two interacting persons for the contact signature estimation, we employ the same architecture of~\cite{fieraru2020three} that has two pathways for each person.

To train $f^\text{Contact}$, $f^\text{CS}$ and $f^\text{sig}$, we use the balanced binary cross entropy (BBCE) loss as follows:
\begin{eqnarray}
    L_{BBCE} = -(\textit{B}\bar{y}\log(y)+(1-\bar{y})\log(1-y))
\end{eqnarray}
where $\bar{y}$, $y$ and \textit{B} denote the ground-truth, estimated value and balance term.
% To train $f^\text{Contact}$, we use the binary cross entropy (BCE) loss as follows:
% \begin{eqnarray}
%     L_\text{Contact}(f^\text{Contact}) = -(\bar{y}\log(y)+(1-\bar{y})\log(1-y))
% \end{eqnarray}
% where $\bar{y}\in\{0,1\}$ denotes the ground-truth label indicating whether there is a contact (\ie 1) or not (\ie 0), and $y\in[0,1]$ denotes the predicted value.

% To train $f^\text{CS}$ and $f^\text{sig}$, we apply the binary cross entropy (BCE) loss on two outputs $\{s_1, s_2\}$, $\mbC$ as follows:
% \begin{eqnarray}
%     L_\text{CS}(f^\text{CS}, f^\text{sig}) &=& -(\sum^{75}_{j=1} \bar{s}_{1j}\log(s_{1j})+(1-\bar{s}_{1j})\log(1-s_{1j}) \nonumber\\
%     &+& \bar{s}_{2j}\log(s_{2j})+(1-\bar{s}_{2j})\log(1-s_{2j}))\nonumber\\ 
%     &-& (\sum^{75 \times 75}_{k=1} \bar{\mbC}_{k}\log(\mbC_{k}) \nonumber\\
%     &+& (1-\bar{\mbC}_{k}) \log(1-\mbC_{k}))
% \end{eqnarray}
% where $\bar{s}_{ij}$, $s_{ij}$ denote the ground-truth and estimated contact segmentation for the $j$-th region of the $i$-th person's mesh, respectively, and $\bar{\mbC}_j$, $\mbC_j$ denote the ground-truth and estimated contact signature for the $j$-th region pair, respectively.

\noindent \textbf{Refinement Stage.} We further refine the scene-space 3D body poses considering people's contacts and their geometry surface. To this end, we select two meshes $\{\mbM_i, \mbM_j\}$ where $i\ne j$, and refine the SMPL pose $\boldsymbol{\theta}$, shape $\boldsymbol{\beta}$ and 3D location $\mbT^\text{3D}$ of two meshes $\mbM$. Note that, if $f^\text{Contact}$ predicts that there is ``no contact'', we skip this stage. We optimize $\mbT^\text{3D}$ of the meshes using the contact loss $L_\text{contact}$ proposed in~\cite{fieraru2020three} as the following formula:
\begin{eqnarray} \mbT^{\text{3D}*}=\arg\min_{\mbT^\text{3D}} L_\text{contact}(\mbT^\text{3D})
\end{eqnarray}
%\sum_{(r_1, r_2) \in \mbC} D(r_1, r_2)
where $L_\text{contact} = \sum^{75}_{r_1=1}\sum^{75}_{r_2=1} \mbC(r_1, r_2)D(r_1,r_2)$. $r_1$ and $r_2$ are 75 regions of each person. $\mbC$ is the estimated contact signature between two regions of interacting persons. It represents 1 if there is contact between $r_1$ and $r_2$, otherwise 0. $D(r_1, r_2)$ denotes the distance between two regions which is formulated as:
\begin{eqnarray}
    D(r_1, r_2) =\sum_{f_1 \in r_1} \min_{f_2 \in r_2} d(f_1, f_2) + \sum_{f_2 \in r_2} \min_{f_1 \in r_1} d(f_1, f_2)
\end{eqnarray}
where $f_1$ and $f_2$ are the faces which belong to two regions $r_1$ and $r_2$, respectively. $d$ is the Euclidean distance between the centers of the two faces $f_1$ and $f_2$. 

\begin{table*}[t]
    \centering
    \begin{tabular}{l|cccccccccc}
        \hline
        Categories & \multicolumn{2}{c}{single} & \multicolumn{2}{c}{occluded single} & \multicolumn{2}{c}{two natural-inter} & \multicolumn{2}{c}{two closely-inter} & \multicolumn{2}{c}{three} \\
        Methods & CD & P2S & CD & P2S & CD & P2S & CD & P2S & CD & P2S\\
        \hline
        \hline
        PIFu~\cite{saito2019pifu} & 2.3240 & 2.5473 & 3.3008 & 3.1365 & 3.6404 & 3.5573 & 4.0191 & 4.2739 & 3.3058 & 3.4899 \\
        PIFuHD~\cite{saito2020pifuhd} & 1.9745 & 2.0258 & 4.2392 & 4.5324 & 3.3252 & 3.0838 & 3.3901 & 3.3945 & 3.2487 & 3.1508 \\
        ICON~\cite{xiu2022icon} & 3.6610 & 4.6765 & 4.6721 & 5.0283 & 5.0958 & 6.1449 & 5.0285 & 6.2282 & 4.1542 & 5.0633 \\
        $\text{DMC$^{\dagger}$}$~\cite{zheng2021deepmulticap} & 2.5878 & 4.0971 & 3.1396 & 4.6999 & 2.5335 & 4.3184 & 3.2481 & 5.3352 & 3.8079 & 5.7033 \\
        $\text{Ours$^{\dagger}$ w/o $\mbN$}$ & 1.4069 & 1.9364 & 1.6574 & \textbf{1.8731} & 1.1996 & 1.9692 & 1.8038 & 2.7844 & 1.7642 & 2.7325 \\
        $\text{Ours$^{\dagger}$}$ & \textbf{1.3730} & \textbf{1.8441} & \textbf{1.6329} & 1.8898 & 1.1933 & 1.9578 & 1.8382 & 2.8614 & 1.7608 & 2.7112 \\
        $\text{Ours$^{\dagger}$ w R.}$ & - & - & - & - & \textbf{1.1723} & \textbf{1.9065} & \textbf{1.3358} & \textbf{2.0669} & \textbf{1.5399} & \textbf{2.2636} \\
        % \hline
        % DMC~\cite{zheng2021deepmulticap} & 1.3961 & 2.2184 & 1.2792 & 2.0194 & 0.9960 & 1.8082 & 1.3258 & 2.3460 & 1.2322 & 2.0885 \\
        % Ours & \textbf{0.7298} & \textbf{0.9609} & \textbf{0.9983} & \textbf{1.4440} & \textbf{0.6718} & \textbf{1.1757} & \textbf{0.8534} & \textbf{1.4475} & \textbf{0.7999} & \textbf{1.2117} \\
        \hline
    \end{tabular}
    \caption{The comparison of our method to the state-of-the-art approaches on MultiHuman~\cite{zheng2021deepmulticap}. $\dagger$ denotes that model uses the estimated SMPL parameters from \cite{cha2022multi}. Ours w/o $\mbN$ denotes the results without using a normal map input $\mbN$. Ours is without refinement based on the contact signature. Ours w R. is our full method. Refinement is not available to ``single'' and ``single occluded'' because they have only one person.}
    % $\ddagger$ denotes that model uses the refined SMPL parameters based on the contact signature. The last two rows use ground-truth SMPL parameters. We mark the best performance in bold for the same setting.
    \vspace{-3mm}
    \label{tab:sota multihuman}
\end{table*}

Finally, we globally refine SMPL pose ($\boldsymbol{\theta}$) and shape ($\boldsymbol{\beta}$) parameters as well as $\mbT^\text{3D}$ via the contact loss $L_\text{contact}$ combined with the penetration loss $L_\text{penet}$~\cite{li2022dig}, Gaussian mixture model prior loss $L_\text{gmm}$~\cite{kolotouros2019learning} and regularizer $L_\text{reg}$ 
as follows:
\begin{eqnarray} 
\boldsymbol{\theta}^*, \boldsymbol{\beta}^*, \mbT^{\text{3D}*}=\arg\min_{\boldsymbol{\theta},\boldsymbol{\beta}, \mbT^\text{3D}} L_\text{opt}(\boldsymbol{\theta}, \boldsymbol{\beta}, \mbT^{\text{3D}})
\end{eqnarray}
where
\begin{eqnarray}
    L_\text{opt}(\boldsymbol{\theta}, \boldsymbol{\beta}, \mbT^{\text{3D}}) &=& \lambda_c L_\text{contact}(\boldsymbol{\theta}, \boldsymbol{\beta}, \mbT^{\text{3D}})\nonumber\\&+&\lambda_p L_\text{penet}(\boldsymbol{\theta}, \boldsymbol{\beta}, \mbT^{\text{3D}})\nonumber\\&+&\lambda_g L_\text{gmm}(\boldsymbol{\theta})\nonumber\\&+&\lambda_r L_\text{reg}(\boldsymbol{\theta}, \boldsymbol{\beta})
\label{eq:regularize loss}
\end{eqnarray}
where $L_\text{penet}=\sum_{v \in V_{1,\text{penet}}} \mbox{SDF}_2 (v) + \sum_{v \in V_{2,\text{penet}}} \mbox{SDF}_1 (v)$ prevents collisions between two people' meshes. The first term of $L_\text{penet}$ means the sum of signed distance fields ($\mbox{SDF}_2$) that measures the distance between the second person's mesh surface and the first person's mesh vertices $V_{1, \text{penet}}$, which are penetrating the surface of the second person's mesh. The second term of $L_\text{penet}$ represents the opposite situation. $\mbox{SDF}_1$ denotes the signed distance fields of the first person's mesh. $V_2$ is the second human vertex set inside the first person's mesh. Gaussian mixture model prior loss $L_\text{gmm}$ reduces the probability of predicting the pose that deviates from distribution and regularizer $L_\text{reg} = \|\boldsymbol{\theta} - \boldsymbol{\theta_\text{init}}\|^2_2+\|\boldsymbol{\beta} - \boldsymbol{\beta_\text{init}}\|^2_2$ ensures the refined SMPL pose and shape parameters do not diverge much from the initial parameters $\{\boldsymbol{\theta_\text{init}}, \boldsymbol{\beta_\text{init}}\}$. $\lambda_c$, $\lambda_p$, $\lambda_r$, $\lambda_g$ balances the weights over the losses.

\noindent \textbf{Updating the Posed Mesh.} After obtaining the refined parameters $\boldsymbol{\theta}^*, \boldsymbol{\beta}^*, \mbT^{\text{3D}*}$, we re-obtain our posed mesh $\mbM^*$ using the forward skinning with new SMPL parameter $\boldsymbol{\beta}^*$ and $\boldsymbol{\theta}^*$. We also translate the re-posed mesh $\mbM^*$ to scene space by adding the $\mbT^\text{3D*}$ to the posed mesh.

\vspace{-1.5mm}
\section{Experiments}
\subsection{Datasets} 
\noindent\textbf{Generation.} We conduct our experiments using THuman2.0~\cite{tao2021function4d}, MultiHuman~\cite{zheng2021deepmulticap}, and 3DPW~\cite{vonMarcard2018} datasets. THuman2.0~\cite{tao2021function4d} is publicly available, and provides 500 high-fidelity 3D human scans with different clothing styles and poses. The 3D scans have the corresponding texture map and SMPL parameters. We use this dataset to train our $f^{z}$, $f^\text{2D}$, and $f^\text{agg}$. MultiHuman~\cite{zheng2021deepmulticap} consists of 150 high-quality 3D human scans with the corresponding texture map. We use it for evaluation. It provides 3D scans with occlusion by humans or objects. Each sample contains 1 to 3 people and each person wears different clothing. This dataset is divided into 5 categories by the level of occlusions and the number of people. In addition, we use 3DPW~\cite{vonMarcard2018} for evaluation. It provides SMPL~\cite{loper2015smpl} pose parameters and clothed mesh. 

\noindent\textbf{Contact estimation.}
We use FlickrCI3D~\cite{fieraru2020three} to train our contact estimators. It provides the paired data of images and the labels of contact validity and its signature. Among them, we filter out noisy ground truth and select 5K paired data for training.

\begin{table}[t]
    \centering
    \begin{tabular}{l|cccccc}
        \hline
 \small        \multirow{2}{*}{Methods} & \multicolumn{2}{c}{Office Call} & 
 \multicolumn{2}{c}{Hug} & \multicolumn{2}{c}{ShakeHands}  \\
 \small        & CD & P2S & CD & P2S & CD & P2S \\
        \hline
        \hline
      \small  PIFu~\cite{saito2019pifu} & 5.26 & 3.61 & 5.77 & 3.81 & 4.60 & 2.80 \\
       \small PIFuHD~\cite{saito2020pifuhd} & 4.93 & 3.19 & 6.30 & 4.43 & 4.30 & 3.10 \\
        \small ICON~\cite{xiu2022icon} & 4.80 & 4.07 & 5.45 & 4.24 & 4.05 & 3.14 \\
         \small DMC$^{\dagger}$~\cite{zheng2021deepmulticap} & 4.30 & 3.68 & 6.67 & 6.30 & 4.36 & 3.95 \\
        \small Ours$^{\dagger}$ w R. & \textbf{2.16} & \textbf{1.79} & \textbf{3.22} & \textbf{2.84} & \textbf{2.93} & \textbf{2.60} \\
        \hline
    \end{tabular}
    \caption{We compare our proposed method with state-of-the-art methods on 3DPW~\cite{vonMarcard2018}. $\dagger$ denotes that model uses the estimated SMPL parameters from \cite{cha2022multi}.}
    \vspace{-3mm}
    \label{tab:sota 3dpw}
    \vspace{-3mm}
\end{table}

\subsection{Metrics}
Following the measuring protocols from existing works~\cite{saito2019pifu,zheng2021deepmulticap}, we use two metrics to measure the reconstruction quality. 

\noindent\textbf{Chamfer Distance (CD).} It measures the bidirectional point distance between the predicted and ground truth meshes based on their closest-set Euclidean distance where the smaller Chamfer distance means more accurate reconstruction results.

\noindent\textbf{Point to Surface (P2S).} It measures the directional surface distance from the prediction to the ground truth based on their closest-set Euclidean distance. The smaller number is the better.

\subsection{Baselines}
We compare our model with other state-of-the-art methods~\cite{saito2019pifu,saito2020pifuhd,xiu2022icon,zheng2019deephuman}. PIFu~\cite{saito2019pifu} and PIFuHD~\cite{saito2020pifuhd} use the pixel-aligned implicit function so that they can generate the pixel-aligned clothed human mesh. 
% PIFuHD is based on a coarse-to-fine framework which can produce more high-fidelity clothed human mesh than PIFu.
ICON~\cite{xiu2022icon} uses the front and back normal map and fuses their features with a signed distance field to generate the detailed clothed human mesh. However, these methods can not reconstruct invisible parts of humans occluded by objects or other people. DeepMultiCap~\cite{zheng2021deepmulticap} fuses multi-view features by the proposed attention module. While it requires multi-view images as inputs for the best performance, we use a single image for the fair comparison. For the quantitative evaluation of PIFu, PIFuHD, and ICON, we manually transform their results from the image to the scene space using the ground-truth meshes: this means that the errors from 3D scene-space pose estimation are factored out from their methods. For qualitative results in Fig~\ref{fig:qualitative comparison}, we visualize their methods without such advantages since predicting the scene-space position of humans is not a part of their algorithms.

\subsection{Results}
\begin{figure*}
    \centering
    \includegraphics[width=1\linewidth]{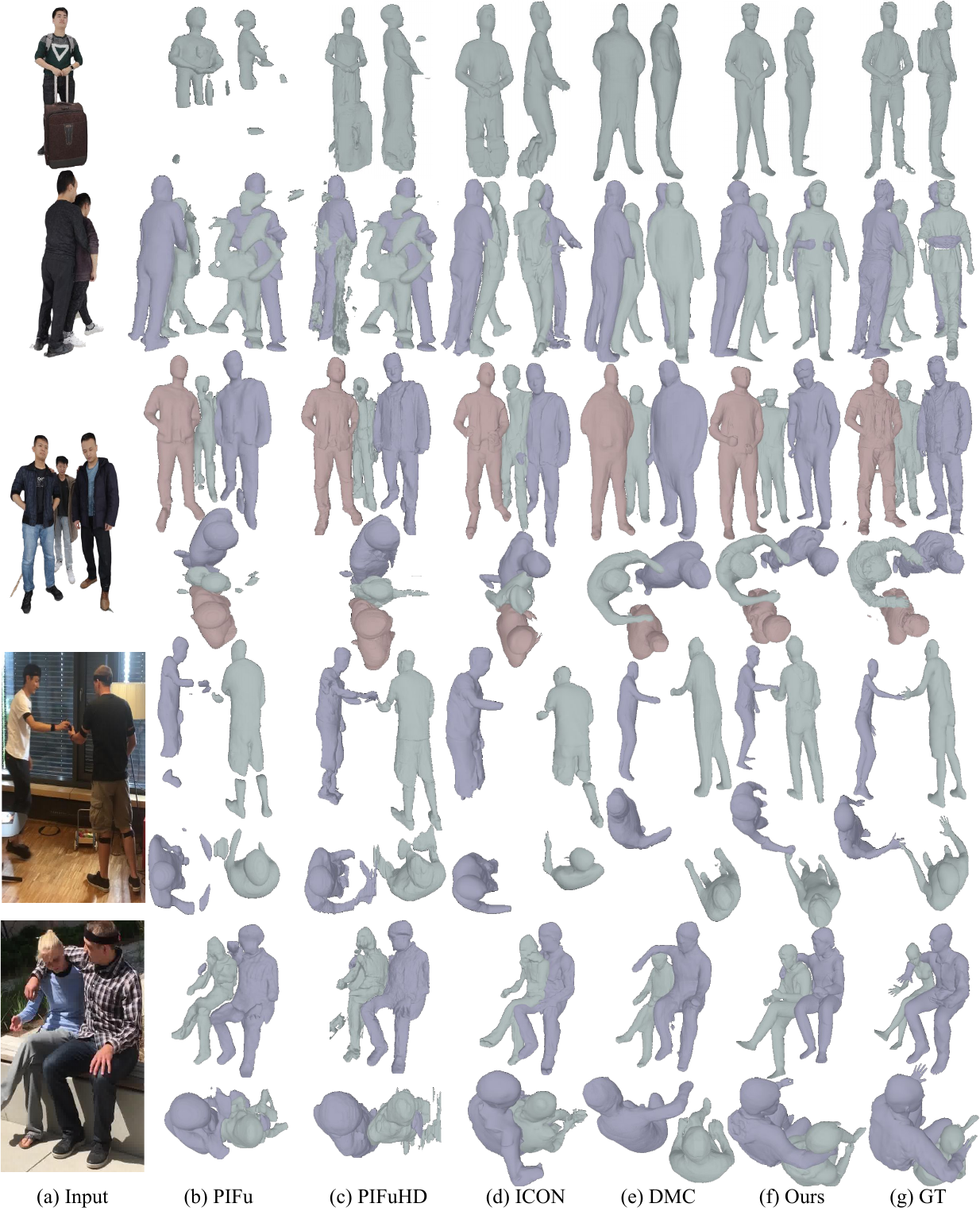}
    \caption{Comparison of qualitative result with state-of-the-art methods, including (b) PIFu~\cite{saito2019pifu}, (c) PIFuHD~\cite{saito2020pifuhd}, (d) ICON~\cite{xiu2022icon}, (e) DMC~\cite{zheng2021deepmulticap}. DMC uses the estimated SMPL parameters from ~\cite{cha2022multi} and our method uses refined SMPL parameters. Our method generates complete human and more detailed human mesh than other methods.}
    \label{fig:qualitative comparison}
\end{figure*}

The comparison results of SOTA and our method are shown in Tab.~\ref{tab:sota multihuman} and Fig~\ref{fig:qualitative comparison}. Overall, our method achieves state-of-the-art performance on MultiHuman~\cite{zheng2021deepmulticap} where in Tab.~\ref{tab:sota multihuman}. Based on Tab.~\ref{tab:sota multihuman} and Fig~\ref{fig:qualitative comparison}, pixel-aligned methods~\cite{saito2019pifu,saito2020pifuhd,xiu2022icon} show the weak reconstruction results for the occluded body parts including a significant amount of artifacts such as missing information. While DeepMultiCap~\cite{zheng2021deepmulticap} produces a mesh that is less noisy compared to other baselines, its geometry reconstruction results are still coarse, and its performance is often affected by the quality of global coordinates estimation of SMPL meshes. Unlike other approaches, our method is able to reconstruct clothed meshes from highly occluded people. Thanks to the global pose refinement with contact signatures, our method is robust to the occlusion, and the active utilization of the upgraded human geometry prior enables the complete and detailed reconstruction of multiple people under occlusion. In addition, Tab.~\ref{tab:sota 3dpw} shows the comparison on in-the-wild data~\cite{vonMarcard2018}. It highlights that our method is generalizable and is able to reconstruct human meshes with details even from unconstrained data. More qualitative results are presented in Supple.

\subsection{Ablation Study}
\begin{figure}
    \centering
    \includegraphics[width=0.99\linewidth]{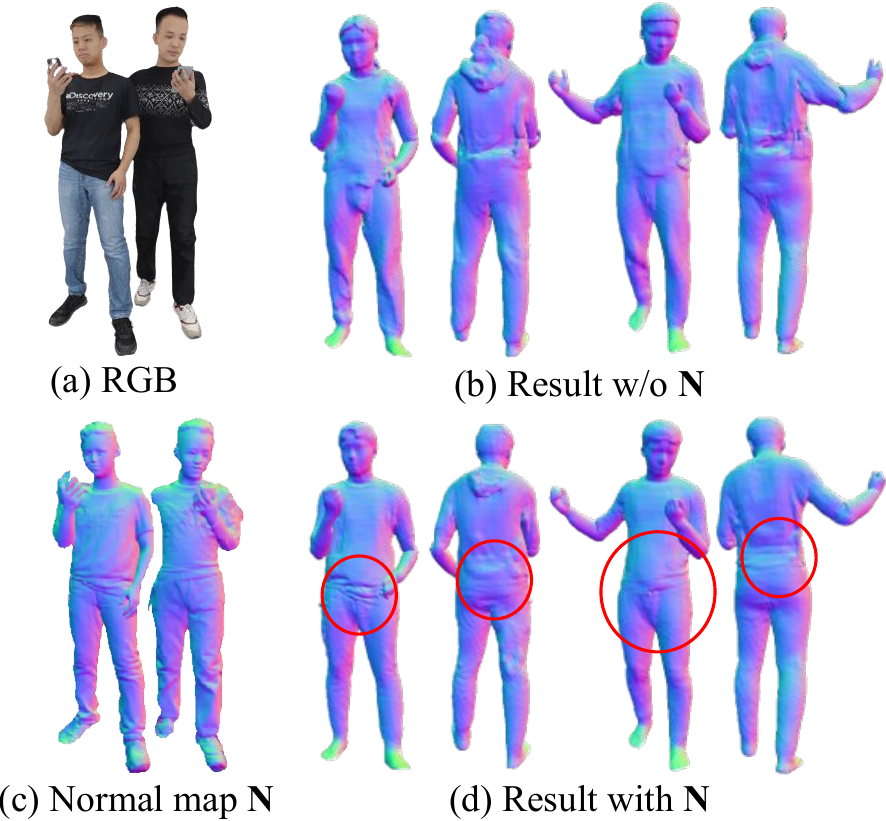}
    \vspace{-3mm}
    \caption{Ablation study on normal map $\mbN$. (a) and (c) are the input RGB image and normal map input, respectively. (b) and (d) show the result without and with normal map, respectively. Normal map $\mbN$ provides more detailed surface information for reconstructing mesh.}
    \vspace{-3mm}
    \label{fig:ablation normal map}
\end{figure}

% \noindent\textbf{Generation.} 
We conduct an ablation study to confirm the effect of our module. In Tab.~\ref{tab:sota multihuman}, we study the performance of our implicit model $f^\text{agg}$ without a normal map input (Ours w/o $\mbN$); and the 3D reconstruction results with the refined body pose and shape (Ours w R.). Refinement with surface contact priors highly improves the performance from the scenes of multi-person (two or three people) under occlusion. Although utilizing a normal map $\mbN$ slightly improves the performance, it can help to express the details of the reconstructed surface as shown in Fig~\ref{fig:ablation normal map}.

% \begin{table}[t]
%     \centering
%     \begin{tabular}{l|cccccc}
%         \hline
%         \small \multirow{2}{*}{} & \multicolumn{2}{c}{Ours w/o $\mbN$} & \multicolumn{2}{c}{Ours w/o r.} & \multicolumn{2}{c}{Ours} \\
%         	 & \footnotesize CD & \footnotesize P2S & \footnotesize CD & \footnotesize P2S & \footnotesize CD & \footnotesize P2S \\
%         \hline
%         \hline
%         \footnotesize single & \footnotesize 1.41 & \footnotesize 1.94 & \footnotesize \textbf{1.37} & \footnotesize \textbf{1.84} & - & - \\
%         \footnotesize single occluded & 1.66 & \textbf{1.87} & \textbf{1.63} & 1.89 & - & - \\
%         \footnotesize two naturally-inter & 1.20 & 1.97 & 1.19 & 1.96 & \textbf{1.17} & \textbf{1.90} \\
%         \footnotesize two closely-inter & 1.80 & 2.78 & 1.84 & 2.86 & \textbf{1.33} & \textbf{2.07} \\
%         \footnotesize three & 1.76 & 2.73 & 1.76 & 2.71 & \textbf{1.54} & \textbf{2.26} \\
%         \hline
%     \end{tabular}
% % Ablation study on the different scenarios. The first column denotes the results without using a normal map input $\mbN$. The second column is the results without using op step. Third is our full method.
%     \caption{Ablation study on the different scenarios. The first column denotes the results without using a normal map input $\mbN$. The second is without refinement based on the contact signature. Third is our full method. Refinement is not available to ``single'' and ``single occluded'' because they have only one person.}
%     \label{tab:ablation}
% \end{table}

\begin{figure}
    \centering
    \includegraphics[width=0.99\linewidth]{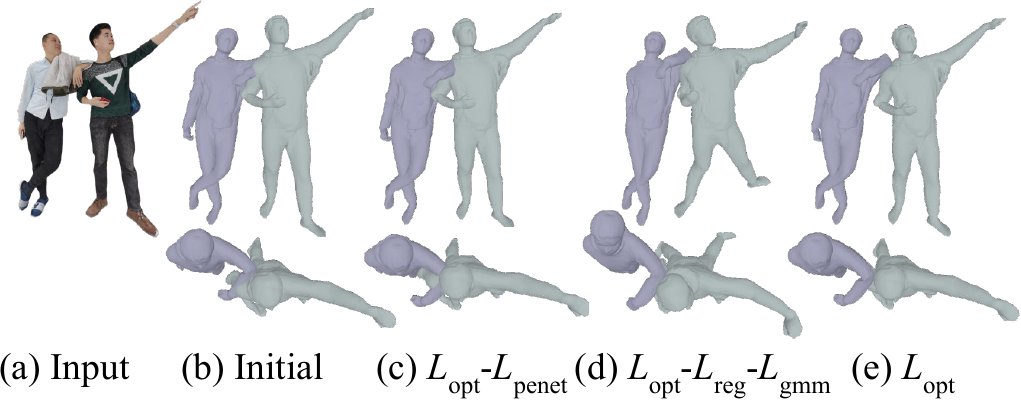}
    \vspace{-3mm}
    \caption{Ablation study on the elements of $L_\text{opt}$. First row shows the results in front view. Second row shows the results in top view. (a) is RGB image input. (b) shows initial posed meshes. (c) shows the results without using $L_\text{penet}$. (d) shows the results without using pose and shape prior loss, $L_\text{reg}$ and $L_\text{gmm}$. (e) shows the results of our full method.}
    \vspace{-5mm}
    \label{fig:contact_loss} 
\end{figure}

We further study the effect of the loss used in our refinement pipeline in $L_\text{opt}$. We use the estimated SMPL parameters as initial SMPL parameters from ~\cite{cha2022multi}. As shown in Fig~\ref{fig:contact_loss}, $L_\text{penet}$ prevents penetration between two people. $L_\text{gmm}$ and $L_\text{reg}$ prevent divergence of SMPL~\cite{loper2015smpl} parameters. Our full $L_\text{opt}$ refines the SMPL parameters and 3D location more realistically based on contact signature.

\begin{figure}
\centering
\includegraphics[width=0.99\linewidth]{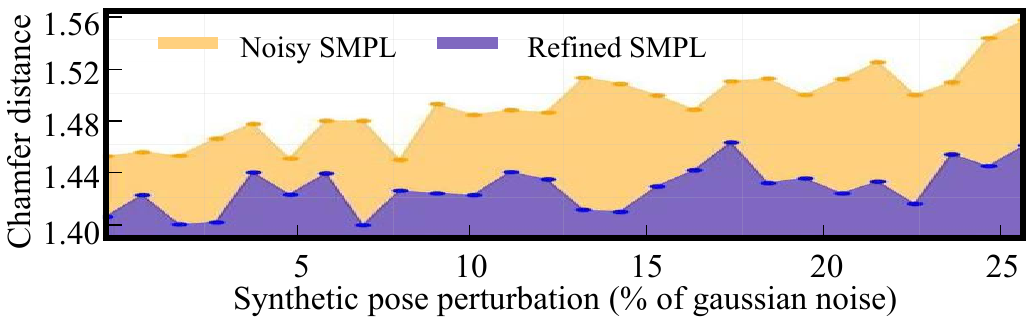}
\vspace{-4mm}
\caption{The reconstruction accuracy (CD) with noisy and refined SMPL parameters. The noisy pose is synthetically generated by adding the fraction of Gaussian noise which perturbs ground-truth SMPL parameters.}
\vspace{-5mm}
\label{fig:pose_perturbation}
\end{figure}

We perform an ablation study in Fig.~\ref{fig:pose_perturbation} where the reconstruction error (\ie chamfer distance) is highly suppressed by our refinement module with surface contact and inter-penetration priors. Our refinement module is robust to pose noise. In the supplemental, we include more ablative and comparative studies.

\section{Conclusion}
We introduce a novel system for scene-space 3D reconstruction of interacting multi-person in clothing from a single image. We address the core challenge of occlusion by utilizing the human priors for complete 3D human geometry and surface contacts. For the geometry, we upgrade the existing generative 3D features by newly designing an implicit network that combines these features and a surface normal map to produce fine-detailed and complete 3D geometry. For the surface contacts, we design an image-based detector that signals the surface contact information between people in 3D. Using these priors, we globally refine 3D body poses to reconstruct accurate and penetration-free human models in scene space. The comparative evaluation and ablation study demonstrate that our method has strong and accurate performance with detailed geometry reconstruction even under heavy occlusion.

\noindent\textbf{Limitation.}
When the occlusion is seriously severe, e.g., only a head part is visible, our method will not be working well due to the complete failure of the initial 3D pose estimation. The geometric diversity is not completely reflective of \textit{in-the-wild} distribution due to the fundamental domain constraints of 3D human geometric prior, e.g., gDNA~\cite{chen2022gdna}. In our future work, we would like to explore the modeling of a better 3D geometric prior and its adaptation to the in-the-wild environment.

%-------------------------------------------------------------------------
\noindent\textbf{Acknowledgements.} This work was supported by IITP grants (No. 2021-0-01778 Development of human image synthesis and discrimination technology below the perceptual threshold 10\%; No. 2020-0-01336 Artificial intelligence graduate school program (UNIST) 10\%; No. 2021-0-02068 Artificial intelligence innovation hub 10\%; No. 2022-0-00264 Comprehensive video understanding and generation with knowledge-based deep logic neural network 20\%) and the NRF grant (No. RS-2023-00252630 10\%), all funded by the Korean government (MSIT). This work was also supported by Korea Institute of Marine Science \& Technology Promotion(KIMST) funded by Ministry of Oceans and Fisheries (RS-2022-KS221674) 20\% and received support from AI Center, CJ Corporation. (20\%).

%%%%%%%%% REFERENCES
{\small
\bibliographystyle{ieee_fullname}
\bibliography{egbib}
}

\twocolumn[{
\begin{center}
\textbf{\Large 3D Reconstruction of Interacting Multi-Person in Clothing from a Single Image\\-Supplementary-\\}
\vspace{16mm}
\end{center}
}]

%%%%%%%%%% Merge with supplemental materials %%%%%%%%%%
%%%%%%%%%% Prefix a "S" to all equations, figures, tables and reset the counter %%%%%%%%%%
\setcounter{equation}{0}
\setcounter{figure}{0}
\setcounter{table}{0}
\setcounter{page}{1}
\makeatletter
\renewcommand{\theequation}{S\arabic{equation}}
\renewcommand{\thefigure}{S\arabic{figure}}
\renewcommand{\thetable}{S\arabic{table}}
% \renewcommand{\bibnumfmt}[1]{[S#1]}
% \renewcommand{\citenumfont}[1]{S#1}
%%%%%%%%%% Prefix a "S" to all equations, figures, tables and reset the counter %%%%%%%%%%

In this supplementary material, we provide the details of the implementation and network architectures used in our pipeline; ablations study on 2D contact networks and 3D reconstruction results; and more results with in-the-wild images. Please also refer to the supplementary video.

\section{Implementation Details}
\vspace{-1.5mm}
We design $f^{z}$ with ResNet encoder~\cite{he2016deep} and one fully-connected layer. $f^\text{3D}$ consists of 3 sequences of fully-connected layers, 3D convolution layers and last one fully-connected layer. The 2D image-aligned feature extractor $f^\text{2D}$ consists of a stacked hourglass~\cite{newell2016stacked} with 1 stack. Aggregation network $f^\text{agg}$ is designed with 5 fully-connected layers. Please see Supple. for the detailed architectures of our networks.
% We design $f^{z}$ with ResNet encoder~\cite{he2016deep} and one fully-connected layer. It extracts 64 dimensions of latent code $z$ from the input image. $f^\text{3D}$ consists of 3 sequences of fully-connected layers, 3D convolution layers and last one fully-connected layer. It outputs $64\times16\times64\times64$-dimensional features $\mbF^\text{3D}$. The 2D image-aligned feature extractor $f^\text{2D}$ consists of a stacked hourglass~\cite{newell2016stacked} with 1 stack. It estimates $16\times512\times512$-dimensional features $\mbF^\text{2D}$. Aggregation network $f^\text{agg}$ is designed with 5 fully-connected layers. Please see Supple. for the detailed architectures of our networks.

% of our networks including $f^\text{Contact}$, $f^\text{CS}$, and $f^\text{sig}$ are presented in the supplemental.}

For training the networks except for $f^\text{3D}$, we use Adam optimizer~\cite{kingma2014adam} with learning rate $1e^{-3}$. To train $f^\text{2D}$ and $f^\text{agg}$, we sample the 2,250 points with 2,000 near the ground-truth mesh surface and 250 uniformly sampled. We train $f^{z}$ for 100 epochs, and $f^\text{2D}$ and $f^\text{agg}$ for 300 epochs. $\lambda_c$, $\lambda_p$, $\lambda_r$, and $\lambda_g$ are set to 500, 10, 100, and 0.001, respectively.

\section{Network Detail}
\subsection{Generation}
\noindent\textbf{Latent Feature Encoder $f^z$} predicts latent feature vector $z_\text{shape}$ from a single RGB image. It consists of ResNet encoder~\cite{he2016deep} and one fully-connected layer. The ResNet encoder outputs $2048$-dim intermediate feature from image $\mbI_{i}\in \mathbb{R}^{256\times256\times3}$. Then, the fully-connected layer extracts $64$-dim latent feature vectors from the intermediate feature. 

\noindent\textbf{3D Voxel Feature Generator $f^\text{3D}$} generates 3D voxel feature $\mbF^\text{3D} \in \mathbb{R}^{64\times16\times64\times64}$ from estimated latent feature vector $z_\text{shape}$. We borrow its architecture from gDNA~\cite{chen2022gdna}. It is composed of three sequences consisting of fully-connected layers and 3D convolution layers, and one fully-connected layer at the end. 

\noindent\textbf{2D Image-aligned Feature Extractor $f^\text{2D}$} extracts 2D image-aligned feature $\mbF^\text{2D} \in \mathbb{R}^{16\times512\times512}$ from normal map $\mbN_{i} \in \mathbb{R}^{1024\times1024\times3}$. We use a stacked hourglass~\cite{newell2016stacked} with 1 stack for its architecture. 

\noindent\textbf{Aggregation Network $f^\text{agg}$} estimates an occupancy value at 3D point $\mbp$, from a concatenated feature [$\mbF^\text{3D}_{\mbp_\text{c}}$, $\mbF^\text{2D}_{\pi(\mbp)}$, $\mbp_\text{c}$]. It consists of 5 fully-connected layers with SoftPlus activation function~\cite{zhao2018novel}. In the fourth layer, we use a skip connection. Its intermediate channel is 256.

\subsection{Contact Estimation}
\noindent\textbf{Contact Discriminator $f^\text{Contact}$} predicts whether two people are contacted to each other or not, from a concatenated image [$\mbI$, $\mbm^\text{part}_1$, $\mbm^\text{part}_2$, $\mbh_1$, $\mbh_2$]. It consists of ResNet encoder~\cite{he2016deep} and two fully-connected layers with ReLU activation function~\cite{nair2010rectified}. We modify the input channel of the first 2D convolution layer in a way that it can use the concatenated images as input.

\noindent\textbf{Contact Segmentation Estimator $f^\text{CS}$} estimates contact segmentation $\mbs \in \mathbb{R}^{75\times1}$ from a concatenated image [$\mbI$, $\mbm^\text{part}_1$, $\mbm^\text{part}_2$, $\mbh_1$, $\mbh_2$]. It consists of ResNet encoder~\cite{he2016deep} and five fully-connected layers with ReLU activation function~\cite{nair2010rectified}. We modify the input channel of the first 2D convolution layer to make it able to use the concatenated images as input.

\noindent\textbf{Contact Signature Estimator $f^\text{sig}$} outputs $\mbF^\text{sig} \in \mathbb{R}^{75\times10}$ to estimate contact signature $\mbC \in \mathbb{R}^{75\times75}$. It consists of two fully-connected layers with ReLU activation function~\cite{nair2010rectified}. We compute contact signature $\mbC$ by $\mbF^\text{sig}_1 \times {\mbF^\text{sig}_2}^T$.

\begin{table}[t]
    \centering
    \begin{tabular}{l|cc}
    \hline
    \multirow{2}{*}{Input type}     & \multicolumn{2}{c}{IoU} \\
     & Segm. & Signature \\
    \hline
    RGB & 0.44 & 0.12 \\
    RGB + $\mbh$ & 0.66 & 0.15 \\
    RGB + $\mbm^\text{part}$ & 0.73 & 0.20 \\
    RGB + $\mbh$ + $\mbm^\text{part}$ & \textbf{0.78} & \textbf{0.31} \\
    RGB + $\mbh$ + part segm.~\cite{lin2020cross} & 0.75 & 0.22 \\
    \hline
    \end{tabular}
    \caption{Comparison the contact estimation performance on MultiHuman dataset~\cite{zheng2021deepmulticap} using different input types. We compare five input types: (1) RGB image only, (2) RGB image and keypoint heatmap $\mbh$, (3) RGB image and semantic part segmentation mask $\mbm^\text{part}$, (4) RGB image, $\mbh$, and $\mbm^\text{part}$, and (5) RGB image, $\mbh$, and part segmentation obtained from ~\cite{lin2020cross}.}
    \label{tab:input type}
\end{table}

\section{Ablation Study}
\subsection{Comparison on 2D Contact Network}
We investigate the performance of the contact estimation using different input types. We use intersection over union (IoU) as our evaluation metric. Tab.~\ref{tab:input type} shows that using additional input types, such as keypoint heatmap $\mbh$ and semantic part segmentation mask $\mbm^\text{part}$, improves the performance of contact estimation compared to using only RGB images. Specifically, adding $\mbh$ to the RGB image improves the performance by 50\% on contact segmentation and 25\% on contact signature, while adding $\mbm^\text{part}$ improves the performance by 65\% on contact segmentation and 67\% on contact signature. Finally, using all three input types results in the best performance, with an improvement of 77\% on contact segmentation and 158\% on contact signature. We also compare with a previous approach~\cite{fieraru2020three} which uses 2D part segmentation. We use the method proposed by Lin~\etal~\cite{lin2020cross} for 2D part segmentation estimation. It improves the performance compared to using RGB and $\mbh$, but it does not outperform the performance of the method which uses our final input type (RGB+$\mbh$+$\mbm^\text{part}$).

\begin{figure}[t]
    \captionsetup[subfigure]{labelformat=empty}
    \centering
    \subfloat[(a) Input]{\includegraphics[width=0.2\linewidth]{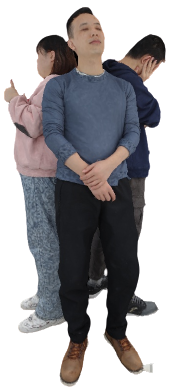}}
    \subfloat[(b) Result]{\includegraphics[width=0.75\linewidth]{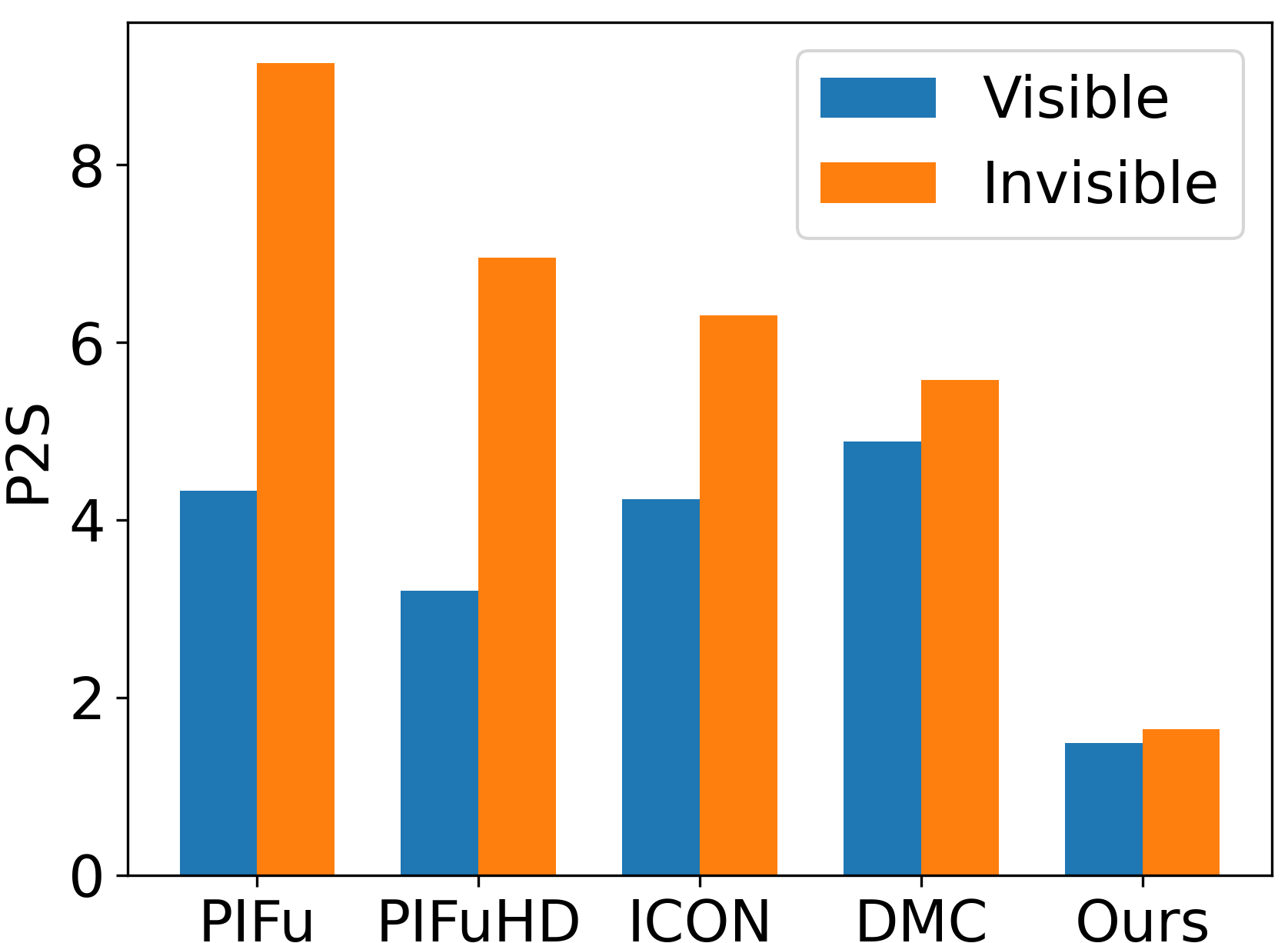}}
    \caption{Comparison on the difference between separate performance measuring on the visible and invisible body parts. (a) shows RGB input image. We compare our method with PIFu~\cite{saito2019pifu}, PIFuHD~\cite{saito2020pifuhd}, ICON~\cite{xiu2022icon}, and DMC~\cite{zheng2021deepmulticap}. ``Visible'' denotes the performance measuring on the visible body parts. ``Invisible'' denotes the performance measuring on the occluded body parts.}
    \label{fig:invisible}
\end{figure}

\begin{table}[t]
    \vspace{-4mm}
    \centering
    \begin{tabular}{cc|cc}
    \hline
    % \multirow{2}{*}{Method}   & \multirow{2}{*}{Single} & Occluded & Two & Two} & \multirow{2}{*}{\scriptsize{Three}} & \multirow{2}{*}{\scriptsize{Avg.}} \\
    % &  & \scriptsize{single} & \scriptsize{natural-inter} & \scriptsize{closely-inter} &  & \\
    % \hline
    % Only $z_{\rm shape}$ & 1.41 & 1.66 & 1.28 & 1.91 & 1.82 & 1.61 \\
    % Ours & \textbf{1.37} & \textbf{1.63} & \textbf{1.17} & \textbf{1.33} & \textbf{1.54}  & \textbf{1.40} \\
    & & \multicolumn{2}{c}{Method} \\
    \cline{3-4}
    & & Baseline & Ours \\
    \hline
    \multirow{5}{*}{Input} & \multicolumn{1}{|c|}{Single} & 1.41 & \textbf{1.37} \\
    & \multicolumn{1}{|c|}{Occluded single} & 1.66 & \textbf{1.63} \\
    & \multicolumn{1}{|c|}{Two natural-inter} & 1.22 & \textbf{1.17} \\
    & \multicolumn{1}{|c|}{Two closely-inter} & 1.91 & \textbf{1.33} \\
    & \multicolumn{1}{|c|}{Three} & 1.82 & \textbf{1.54} \\
    \hline
    \end{tabular}
    \caption{Comparison of our method with the baseline that predicts $z_\text{shape}$ on MultiHuman dataset~\cite{zheng2021deepmulticap}, based on Chamfer Distance metric.}
    \label{tab:baseline}
\end{table}

\subsection{Comparison on Baseline Predicting $z_\text{shape}$}
We compare our full model to the baseline, $f^{z}$, that predicts $z_\text{shape}$. Predicting only $fz_\text{shape}$ cannot outperform the performance of original gDNA~\cite{chen2022gdna}. With Chamfer Distance error in Tab.~\ref{tab:baseline}, our proposed networks, such as 2D image-aligned feature extractor and aggregation network, as well as the refinement module, enhance the level of detail and global coherence of the human mesh. 

\begin{table}[t]
    \centering
    \begin{tabular}{l|c|cc}
    \hline
    Method & Scenario & initial & refined \\ 
    \hline
    DMC  & \multirow{2}{*}{Two natural-inter} & 2.53 & 2.50 \\
    Ours & & 1.19 & \textbf{1.17} \\
    \hline
    DMC  & \multirow{2}{*}{Two close-inter} & 3.24 & 3.10 \\
    Ours & & 1.84 & \textbf{1.34} \\
    \hline
    DMC  & \multirow{2}{*}{Three people} & 3.81 & 3.76 \\
    Ours & & 1.76 & \textbf{1.54} \\
    \hline
    \end{tabular}
    \caption{Comparison with DMC on MultiHuman dataset~\cite{zheng2021deepmulticap}, based on Chamfer Distance metric. `initial' and `refined' denote initial SMPL from \cite{cha2022multi} and SMPL refined by our refinement module.}
    \label{tab:refinement module}
\end{table}

\subsection{Effectiveness of Refinement Module}
Based on the comparison with DMC in Tab.~\ref{tab:refinement module}, we highlight that our refinement module is compatible with any off-the-shelf mesh reconstruction method to improve its accuracy. Performance improvement through the refinement module is observed for all methods and scenarios.

\begin{figure}[t]
\centering
\includegraphics[width=1\linewidth]{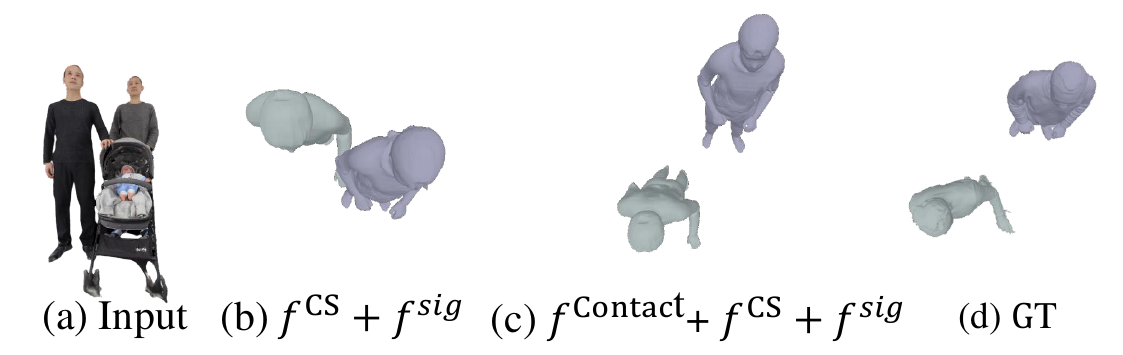}
\caption{The 3D reconstruction results without and with $f^\text{Contact}$.}
\label{fig:classification}
\end{figure}

\subsection{Performance on the Occluded and Visible Body Parts}
Fig.~\ref{fig:invisible} shows the difference between separate performance measuring on the visible and invisible body parts for a representative example. We use point-to-surface (P2S) as our evaluation metric. ``Visible'' denotes the P2S measuring on the visible body parts. ``Invisible'' denotes the P2S measuring on the invisible body parts occluded by other people. While the state-of-the-art methods exhibit a significant performance gap between the visible and invisible parts, our method demonstrates a small difference in performance between the two areas. Furthermore, our method outperforms all the compared approaches in terms of overall performance.

\section{More Qualitative Results}
\begin{figure*}[t]
    \centering
    \includegraphics[width=1\linewidth]{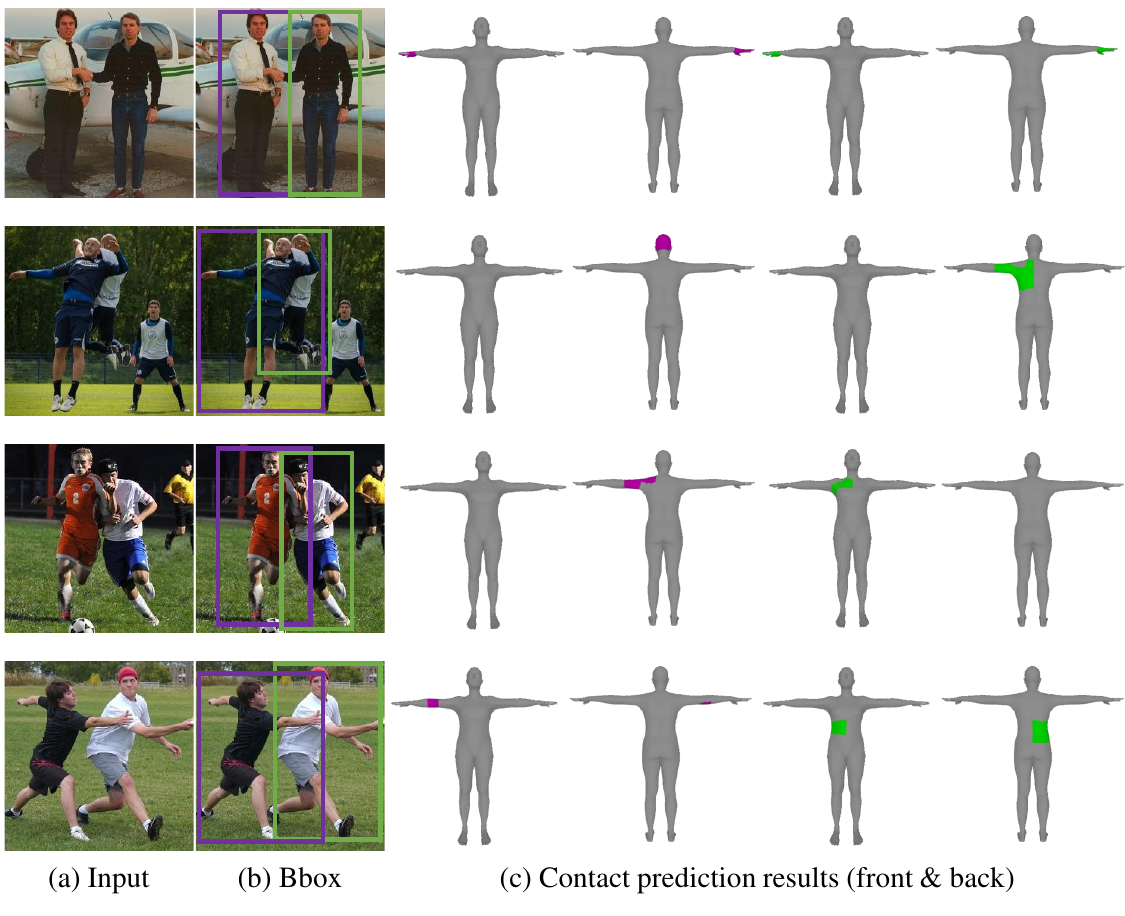}
    \vspace{-8mm}
    \caption{We show the contact prediction results for in-the-wild images from our contact estimator. (a) Input for contact estimator, (b) the image is marked with boxes of the same color corresponding to the contact prediction results, and (c) the contact prediction results on the front and back of the mesh.}
    \label{fig:contact}
\end{figure*}

\begin{figure*}[t]
    \centering
    \includegraphics[width=1\linewidth]{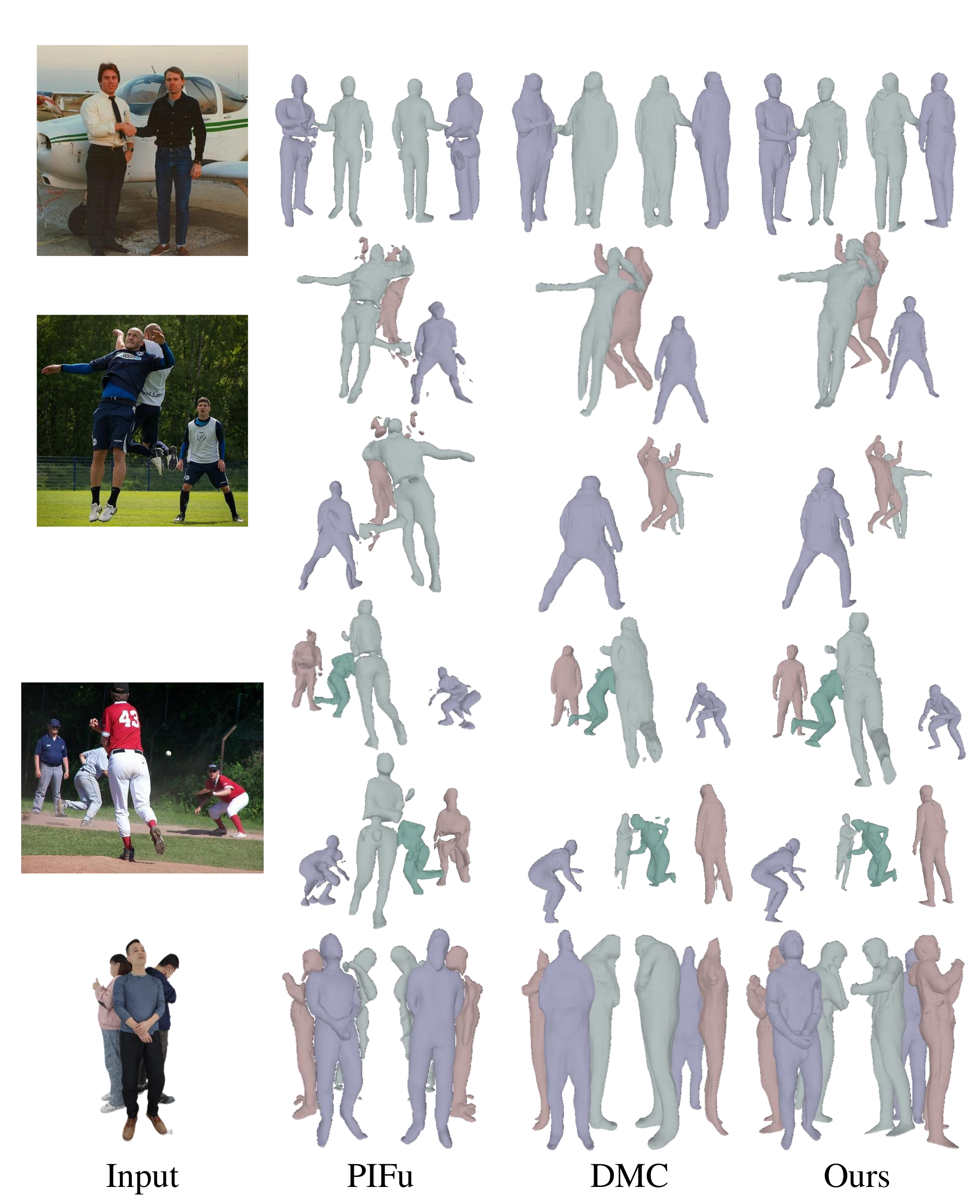}
    \caption{Results comparison on in-the-wild images and MultiHuman image~\cite{zheng2021deepmulticap}. We compare our method with PIFu~\cite{saito2019pifu} and DMC~\cite{zheng2021deepmulticap}. We visualize the results in front-view and back-view.}
    \label{fig:inthewild}
\end{figure*}

\begin{figure*}[t]
    \centering
    \includegraphics[width=1\linewidth]{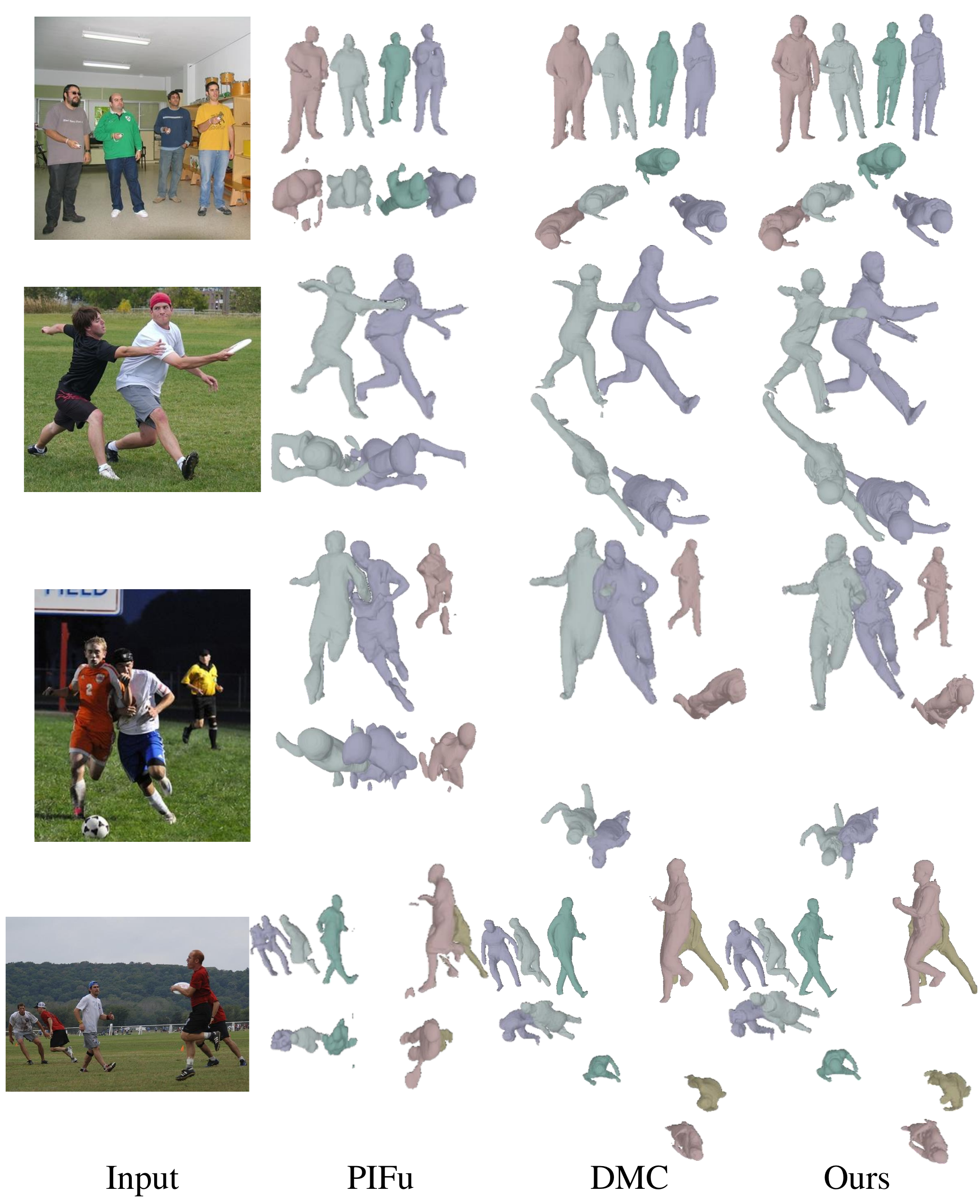}
    \caption{Results comparison on in-the-wild images. We compare our method with PIFu~\cite{saito2019pifu} and DMC~\cite{zheng2021deepmulticap}. We visualize the results in front-view and top-view.}
    \label{fig:inthewild2}
\end{figure*}

We visualize our contact prediction results on the front and back of the mesh as shown in Fig.~\ref{fig:contact}. The columns show the contact estimator input, bounding boxes colored according to the contact prediction results, and contact prediction results on the mesh, respectively. (Note that bounding boxes are not estimated results, and just make it easier to find individuals in the input image who correspond to the contact result.) The contact estimator predict the contact region such as hand, head, back, arm, etc. Sometimes, the contact prediction results are not accurate due to depth ambiguity, as shown in the fourth row. The reconstruction results for each image in Fig.~\ref{fig:contact} are presented in Figs.~\ref{fig:inthewild} and ~\ref{fig:inthewild2}.

We compare our method with PIFu~\cite{saito2019pifu} and DMC~\cite{zheng2021deepmulticap} on in-the-wild images and MultiHuman image~\cite{zheng2021deepmulticap} in Fig.~\ref{fig:inthewild} and Fig.~\ref{fig:inthewild2}. We use COCO dataset~\cite{lin2014microsoft} to compare the qualitative results on in-the-wild images. In Fig.~\ref{fig:inthewild}, the first to third rows of input images are in-the-wild images, while the fourth row shows result on a MultiHuman image. The first and fourth rows show the front-view and back-view results side by side. The second and third rows show the front-view and back-view results vertically. In Fig.~\ref{fig:inthewild2}, all results are presented in a vertical format, displaying both front-view and top-view results for each row on in-the-wild images. PIFu does not consider the depth location, and therefore, it can not represent the perspective distance(\ie large for near and small for far). In addition, it can not reconstruct the 3D geometry for the invisible parts. DMC generates a relatively complete human mesh when compared to PIFu. Nevertheless, the meshes reconstructed by DMC are coarse and lack some parts. Conversely, our method reconstructs the complete and detailed human mesh, even in the presence of occlusions. 

\section{Inference speed}
For comparison, PIFu~\cite{saito2019pifu} takes 5 seconds, and DMC~\cite{zheng2021deepmulticap} takes 41 seconds to generate a clothed human mesh. In contrast, our model, excluding the contact-based refinement, requires 6 seconds to create a clothed human mesh using the Marching Cubes algorithm~\cite{lorensen1987marching} from an occupancy field. Additionally, our refinement module takes 58 seconds to refine two meshes based on contacts. These inferences were performed on an RTX Titan GPU.

\section{Training data}
\begin{figure}[t]
\centering
\begin{subfigure}{0.4\linewidth}
    \includegraphics[width=\linewidth]{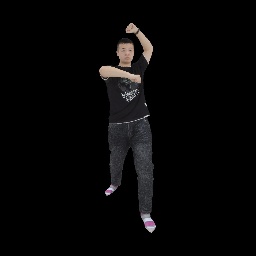}
\end{subfigure}
\begin{subfigure}{0.4\linewidth}
    \includegraphics[width=\linewidth]{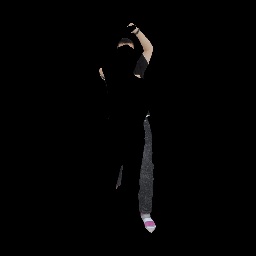}
\end{subfigure}
\caption{Training data. Left shows an original image and right shows a masked input image.}
\label{fig:train_image}
\end{figure}

We train our network using THuman2.0~\cite{tao2021function4d} dataset. For the training images, we augment the images by masking them with the segmentation masks of humans to simulate occlusion, thereby enabling the model to learn how to handle partially occluded humans. This is illustrated in Fig.~\ref{fig:train_image}. Furthermore, we use color-jitter for data augmentation.

\section{Failure Case}
\begin{figure}[t]
\centering
\includegraphics[width=1\linewidth]{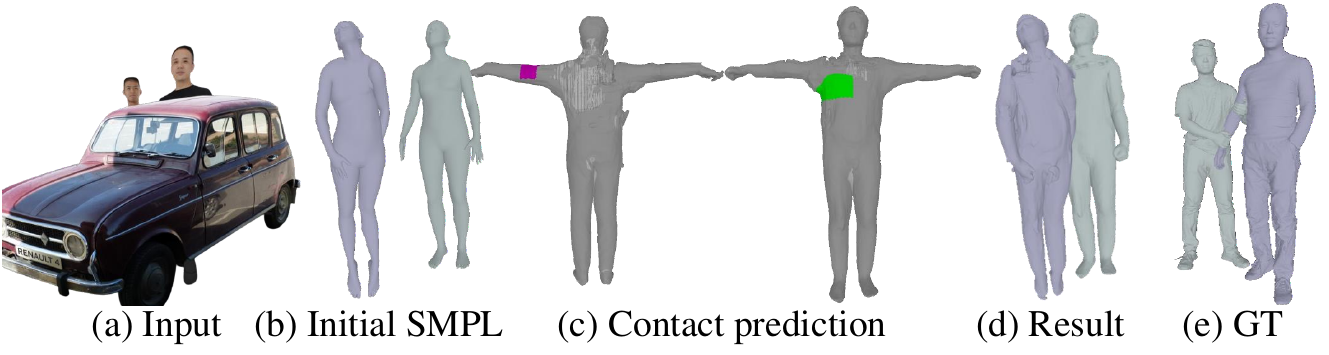}
\caption{Failure case: a strong occlusion leads to the failure of contact prediction (left arm and right chest) and inaccurate 3D reconstruction results. Left to right: input, initial SMPL, contact prediction result, reconstruction result, and ground-truth.}
\label{fig:fail}
\end{figure}

In Fig.~\ref{fig:fail}, our method demonstrates weak performance in the presence of strong occlusion which leads to the failure of contact estimation and inaccurate 3D mesh reconstruction.

\end{document}